\documentclass[11pt,a4paper]{article}
\usepackage[hyperref]{naaclhlt2018}
\usepackage{times}
\usepackage{latexsym}
\usepackage{url}
\usepackage{graphicx}
\usepackage{subcaption}
\usepackage[normalem]{ulem} % for \sout to strike out
\usepackage{enumitem}
\usepackage{multirow} %for merge rows in a table

\aclfinalcopy % Uncomment this line for the final submission
\def\aclpaperid{28} %  Enter the acl Paper ID here

\title{Annotation and Classification of Sentence-level Revision Improvement}

\author{Tazin Afrin \\
  University of Pittsburgh\\
  Pittsburgh, PA 15260\\
  {\tt tazinafrin@cs.pitt.edu} \\\And
  Diane Litman \\
  University of Pittsburgh\\
  Pittsburgh, PA 15260\\
  {\tt litman@cs.pitt.edu} \\}
\date{}
\begin{document}
\maketitle

\begin{abstract}

Studies of writing revisions rarely focus on revision quality. To address this issue, we introduce a corpus of between-draft revisions of student argumentative essays, annotated as to whether each revision improves essay quality.
%We calculate annotation reliability under two different Amazon Mechanical Turk data collection conditions and present insights gained from the results.  
We demonstrate a potential usage of our annotations 
%in facilitating research in writing analysis 
by developing a machine learning model to predict revision improvement. %Our model shows significantly better performance compared to the baseline.
With the goal of expanding training data, we also extract revisions from a dataset edited by expert proofreaders. Our results indicate that blending expert and non-expert revisions increases model performance, with expert data particularly important for predicting low-quality revisions. %Our study shows future direction for revision improvement prediction with promising result.

\end{abstract}
%%%%%%%%%%%%%%%%%%%%%%%%%%%%%%%%%%%%%%%%%%%%%%%%%%%%%%%%%%%%%%%%%%%%%%%%%%%%%%%%
\section{Introduction}
\label{sec: introduction}
Supporting student revision behavior is an important area of writing-related natural language processing (NLP) research. 
%Revision is an essential stage of writing to organize our thoughts towards an enriched version of the text. 
While revision is particularly effective in response to detailed feedback by an instructor~\cite{paulus1999}, human writing evaluation  is time-consuming.
To help students improve their writing skills, various writing assistant tools have thus been developed~\cite{elireview,turnitin,ets-writing-mentor,grammarly}.
%Tazin-need to have better citations, either papers or URLS
%~\newcite{elireview} introduces a peer learning environment via feedback,~\newcite{turnitin,ets-writing-mentor} assists in argumentative writing and coherent structure of the essay,~\newcite{grammarly} helps with proofreading and word choices. 
While these tools offer instant feedback on a particular writing draft,
%essay about writing structure, 
they typically fail to explicitly compare revisions between drafts.

Our long term goal is to build a system for supporting students in revising argumentative essays, where the system automatically compares multiple drafts and provides useful feedback (e.g., informing students whether their revisions are improving the essay). 
%Providing such feedback is a challenging task in natural language processing (NLP).
One step towards this goal is the development of  a machine-learning model
to automatically analyze revision improvement. Specifically, given only two sentences - original and revised, our current goal is to predict if a revised sentence is better than the original.% or not.
%based on the characteristics of the text that was changed. 
%We want this corpus to serve as a gold standard as a revision improvement prediction component for use in an automatic revision analysis system.
%We consider two corpora with paired sentences for predicting revision improvement. Essays in our first corpus are revised by non-expert writers and the second corpus is revised by expert proofreaders. 

In this paper, we focus on predicting revision improvement using non-expert (i.e., student) writing data. We first introduce a corpus of paired original and revised sentences that has been newly annotated as to whether each revision made the original sentence better or not. The revisions are a subset of those in the freely available ArgRewrite corpus~\cite{zhang2017hh}, with improvement annotated using standard rubric criteria for evaluating student argumentative writing.
By adapting NLP features used in previous revision classification tasks, we then develop  a prediction model that outperforms baselines, even though the size of our non-expert revision corpus is small. %Estimating how accurately a predictive model will perform is difficult with such a small dataset.
Hence, we explore extracting paired revisions from an expert edited dataset to increase training data. The expert revisions are a subset of those in the freely available Automated Evaluation of Scientific Writing (AESW) corpus~\cite{daudaravicius2016bv}.
%We adapt NLP features used in previous revision classification tasks and 
Our experiments show that with proper sampling, combining expert and non-expert revisions can improve prediction performance, particularly for low-quality revisions. %We believe, our study is a step towards addressing the technical need of automatic revision quality feedback as well as increasing our understanding of revisions.

%%%%%%%%%%%%%%%%%%%%%%%%%%%%%%%%%%%%%%%%%%%%%%%%%%%%%%%%%%%%%%%%%%%%%%%%%%%%%%%%
\section{Related Work}
\label{sec: related work}
%Prior NLP writing research has largely analyzed single drafts of texts with respect to many linguistic dimensions,  e.g., discourse analysis ~\cite{burstein2003m,falakmasir2014a}, argument mining~\cite{persing2016n,ghosh2014m}, fine-grained error detection~\cite{leacock2010c,grammarly}, etc.  Our work, in contrast,  focuses on analyzing changes between multiple versions of a text, i.e., on revisions. 
Prior NLP revision analysis work has developed methods for identifying pairs of original and revised textual units in both Wikipedia articles and student essays, as well as for classifying such pairs with respect to schemas of coarse (e.g., syntactic versus semantic) and fine-grained  (e.g., lexical vs. grammatical syntactic changes) revision purposes~\cite{bronner2012m,daxenberger2012g,zhang2015l,yang2017hkh}. 
For example, the ArgRewrite corpus~\cite{zhang2017hh} was introduced with the goal to facilitate argumentative revision analysis and automatic revision purpose classification. However, purpose classification does not address revision quality. For example, 
%added evidence can but might not link to a claim, while 
a spelling change can both fix as well as introduce an error, while lexical changes can both enhance or reduce fluency. 
On the other hand, while some work has focused on correction detection in revision~\cite{dahlmeier2012n,xue2014h,felice2016bb},  such work has typically been  limited to grammatical error detection. The AESW shared task of identifying sentences in need of correction~\cite{daudaravicius2016bv} goes beyond just grammatical errors, but the original task does not compare multiple versions of text, and also focuses on scientific writing.

%On the other hand, the goal of AESW shared task~\cite{daudaravicius2016bv} was to promote writing evaluation tool by finding good characteristics (i.e. quality) of scientific writing. However, they do not consider multiple versions of text.

In contrast, \newcite{tan2014l} created a dataset of paired revised sentences in academic writing annotated as to whether one sentence was stronger or weaker than the other. Their work directly sheds light on annotating sentence revision quality in terms of statement strength. However, their corpus focuses on the abstracts and introductions of ArXiv papers.
%, which may not generalize well to other types of writing.
Building on their annotation methodology, we consider paired sentences as our revision unit, but 1) annotate revision quality in terms of argumentative writing criteria, 2) use a corpus of revisions from non-expert student argumentative essays, and 3) move beyond annotation to automatic revision quality classification.

%%%%%%%%%%%%%%%%%%%%%%%%%%%%%%%%%%%%%%%%%%%%%%%%%%%%%%%%%%%%%%%%%%%%%%%%%%%%%%%%
\section{Corpora of Revised Sentence Pairs}
\label{sec: data}
%We consider two different corpus with paired original and revised sentences for predicting sentence-level revision improvement. Our first corpus is Automated Evaluation of Scientific Writing (AESW)~\citep{daudaravicius2016bv} dataset revised by expert writers. This data was introduced to facilitate automatic evaluation tool to assist authors in writing scientific papers. The second corpus  is a subset of those in the freely available ArgRewrite corpus~\cite{zhang2017hh}, where the essays are revised by non-expert writers. We annotated the sentence pairs as to whether the revision made the original sentence better or not, with respect to standard rubrics for evaluating student argumentative writing.

%%%%%%%%%%%%%%%%% Table %%%%%%%%%%%%%%%%%%%%%%%%%
\begin{table*}[!htb]
\centering
\resizebox{2\columnwidth}{!}{
\begin{tabular}{|l|p{0.90\columnwidth}|p{0.90\columnwidth}|l|}
    \hline
    &\textbf{Original Sentence ($S1$)} & \textbf{Revised Sentence ($S2$)} & \textbf{Label} \\ 
    \hline
    \hline
    %659 7-0 7-0
    1&
    The world has \textit{experienced various changes throughout its lifetime}. &
    The world has \textit{been defined by its revolutions - the most recent one being technological}. & Better\\
    \hline
    %112 6-1 2-5
    %Electronic communication cannot \textit{enrich} emotional and physical interactions \textit{between persons}. &
    %Electronic communication cannot \textit{support} emotional and physical interactions. & NotBetter \\
    %\hline
    %35 2-5 1-6
    2&
    Technology is changing the \textit{world, and in particular the} way we communicate. &
 	Technology is changing the way we communicate. & NotBetter \\
 	\hline    
    %637 5-2 6-1
    %Electronic communication \textit{also doesn't hinder making new connections}. &
    %\textit{New connections are easier to make with} electronic communication.  & Better \\ 
    %\hline
    
    % 139 3-4 7-0
    3&
	...Susan says by to Shelly on the 125th St... &
	...Susan says by\textit{e} to Shelly on the 125th St... & Better \\
    \hline \hline
    
    %AESW revisions
    4&
    This is numerically expensive but leads to proper results. & 
    This is numerically expensive\textit{,} but leads to proper results. & 
    Better \\
    \hline
    5&
	Section 2 \textit{formulates and solves} the balance equations. &
	The balance equations \textit{are formulated and solved in} Section 2. &
	Better\\
	\hline
    
\end{tabular}
}
\caption{Example annotated revisions from ArgRewrite (1,2,3) and AESW (4,5). The label is calculated using majority voting (out of $7$ annotators) 
%of $Better$ vs $NotBetter$ annotations 
for ArgRewrite and using expert proofreading edits for AESW.}
\label{table:revision examples}
\end{table*}

%%%%%%%%%%%%%%%%% Table %%%%%%%%%%%%%%%%%%%%%%%%%
\begin{table*}[!htb]
\centering
\resizebox{1.9\columnwidth}{!}{%
\begin{tabular}{|l|c|c|c|c|}
    \hline
\textbf{Data} 						&	\textbf{\#Revisions}  	& \textbf{\#Better} & \textbf{\#NotBetter}	&	\textbf{Fleiss's Kappa ($\kappa$)}	\\  \hline
\hline
All							&	940(100\%) 				& 784(83.4\%) 		& 156(16.6\%)							&	0.201(Slight)	\\ \hline
Majority$\geq5$	&	748(79.6\%) 				& 658(88.0\%) 		& 90(12.0\%)				&	0.263(Fair)	\\ \hline

\end{tabular}
}
\caption{Number of revisions, number of $Better$ and $NotBetter$,  and Fleiss's kappa ($\kappa$) per increasing majority voting (out of $7$ annotators).
Percentage of revisions are shown in parenthesis.}
%\vspace{-1em}
\label{table:inter-annotator agreement}
\end{table*}

\subsection{Annotating ArgRewrite}
\label{sec: argrewrite dataset}
%Each sentence-level revision was annotated with respect to a purpose schema (e.g., add evidence, change spelling)~\cite{zhang2017hh}~\footnote{http://argrewrite.cs.pitt.edu}.
The revisions that we annotated for improvement in quality are a subset of the freely available ArgRewrite revision corpus~\cite{zhang2017hh}\footnote{http://argrewrite.cs.pitt.edu}. This corpus was created by extracting revisions from three drafts of argumentative essays written by $60$ non-expert writers in response to a prompt\footnote{Prompt shown in supplemental files.}.
%Tazin, I don't think you can keep this in the paper, you were supposed to check if it needs to be be uploaded as a separate file and if so you need to fix this.
Essay drafts were first manually aligned at the sentence level based on semantic similarity.
%, with added or deleted sentences aligned to {\it null}. 
Non-identical aligned sentences (e.g., modified, added and deleted sentences) were then extracted as the revisions.
Our work uses only the  940 modification revisions,
%where neither aligned sentence was {\it null}. 
%representing $53\%$ of all ArgRewrite revisions. 
%We excluded  adds and deletes 
as our annotation does not yet consider a sentence's context in its paragraph.

%\subsubsection{Annotating Revision Improvement}
%\label{annotating revision improvement}

% \subsection{Annotation Labels}
We annotated ArgRewrite revisions for improvement using the labels \textit{Better} or \textit{NotBetter}. \textit{Better} is used when the modification yields an improved sentence from the perspective of argumentative writing, while \textit{NotBetter} is used when the modification either makes the sentence worse or does not have any significant effect. Binary labeling enables us to clearly determine a gold-standard using majority voting  with an odd number of annotators.
%\footnote{Three labels were used in a pilot study - $Better$, $Worse$, and $Same$, but it was difficult to decide on the gold-standard as often two labels had an equal number of majority votes.} 
Binary labels should also suffice for our long term goal of triggering tutoring in a writing assistant (e.g., when the label is \textit{NotBetter}).

%\paragraph{Labeling Instructions}
Inspired by~\newcite{tan2014l}, %which for labeling statement strength. 
our annotation instructions included explanatory guidelines along with example annotated sentence pairs. 
The guidelines were crafted to describe improvement in terms of typical argumentative writing criteria. %(e.g., content changes such as modifying an argument's claim, surface changes such as modifying grammar or spelling)~\cite{zhang2015l}. 
We depend on annotators' judgment  for cases not covered by the guidelines. According 
%Tazin - same comment about supplemental material file
to the guidelines\footnote{The  guidelines can be found in supplemental files.}, a revised sentence $S2$ is better than the original sentence $S1$ when: 
%\begin{itemize}[noitemsep]
%    \item 
(1) $S2$ provides more information that strengthens the idea/major claim in $S1$;
 %   \item
 (2) $S2$ provides more evidence/justification for some aspects of $S1$;
 %   \item 
 (3) $S2$ is more precise than $S1$;
%    \item 
(4) $S2$ is easier to understand compared to $S1$ because it is fluent, well-structured, and has no unnecessary words; and
%    \item 
(5)
$S2$ is grammatically correct and has no spelling mistakes.
%\end{itemize}

To provide context, 
%to be annotated, 
annotators were told that the data was taken from student argumentative essays about electronic communications. We also let the annotators know the identity of the original and revised sentences ($S1$ and $S2$, respectively). Although this may introduce an annotation bias, it mimics feedback practice where instructors know which are the original versus revised sentences. 
%\paragraph{Annotation Reliability}
%\label{section:annotation reliability}

We collected $7$ labels along with explanatory comments for each of the $940$ revisions using Amazon Mechanical Turk (AMT).
Table~\ref{table:revision examples} shows examples (1, 2, and 3) of original and revised ArgRewrite sentences with their majority-annotated labels. The first revision clarifies a claim of the essay, the second  removes some information and is less precise, while the third fixes a spelling mistake. As shown in Table~\ref{table:inter-annotator agreement}, for all $940$ revisions, our annotation has %acceptable percent agreement\footnote{Percent agreement is calculated as $\frac{1}{N}\sum_{i=1}^{N} \frac{{{l_{1i}}\choose{2}}+{{l_{2i}}\choose{2}}}{{{L_i}\choose{2}}}$, where $N$ is number of samples, $l_{1i}$ is number of vote for label $Better$, $l_{2i}$ is number of vote for label $NotBetter$, and $L_i = l_{1i} + l_{2i}$} and 
slight agreement~\cite{landis77} using Fleiss's kappa~\cite{fleiss1971}. 
If we only consider revisions where at least 5 out of the 7 annotators chose the same label (majority $\geq 5$), the kappa values increase to fair agreement, $0.263$. 
%This is similar to the statement strength reliability of 
\newcite{tan2014l} achieve fair agreement (Fleiss's kappa of $0.242$) with 9 annotators labeling 500 sentence pairs for statement strength. 

\subsection{Sampling AESW}
\label{sec: aesw dataset}
The Automated Evaluation of Scientific Writing (AESW)~\cite{daudaravicius2016bv} shared task was to predict whether a sentence needed editing or not.
%with a desire to find good and acceptable edits. 
Professional proof-readers edited sentences to correct issues ranging from grammatical errors to stylistic problems, intuitively yielding `Better' sentences. Therefore, we can use the AESW edit information to create an automatically annotated corpus for revision improvement. In addition, by randomly flipping sentences we can include `NotBetter' labels in the corpus.
%To be consistent with the revision types in ArgRewrite data we ignore all sentence with any mathematical notations and only consider plain textual revisions. We randomly flipped original and revised sentences to balance the labels in the data.

%\subsubsection{AESW Data Sampling:} 
The AESW dataset was created from different scientific writing genres (e.g. Mathematics, Astrophysics) with placeholders for anonymization. We use two random samples of 5000 AESW revisions for the experiments in Section~\ref{sec: experiments}. ``AESW all" samples revisions from all scientific genres, while  ``AESW plaintext"
ignores sentences containing placeholders (e.g. MATH, MATHDISP) to make the data more similar to ArgRewrite. %We observe very slow and random decrement in the model performance with increasing AESW data. 
Table~\ref{table:revision examples} shows two example (4 and 5) AESW revisions.% with their labels. 
%The fourth revision corrects uses a punctuation, while the fifth revision re-organizes the original sentence and is more fluent.

%%%%%%%%%%%%%%%%%%%%%%%%%%%%%%%%%%%%%%%%%%%%%%%%%%%%%%%%%%%%%%%%%%%%%%%%%%%%%%%%
\section{Features for Classification}
\label{sec: features}
%While we plan to predict revision improvement, different revision classification tasks are also relevant. 
We adapt many features from prior studies predicting revision purposes~\cite{adler2011dm,javanmardi2011ml,bronner2012m,daxenbergerG13g,zhang2015l,remse2016msAESW} as well as introduce new features tailored to predicting improvement. 
%We follow prior work for argumentative revision classification~\cite{zhang2015l} to extract n-gram, textual, and language features. We also explore parse tree features used by ~\newcite{remse2016msAESW}.
%Besides we use off-the-shelf tools to compare sentence quality. We compare our model performance with majority baseline.

% \paragraph{Majority Baseline:} 
%Our non-expert ArgRewrite dataset is skewed. $83.4\%$ revisions are annotated as $Better$ and $16.6\%$ revisions are annotated as $NotBetter$. That is why we consider majority vote as our baseline instead of considering count of n-gram used in previous works~\cite{daxenbergerG13g,zhang2015l}. Moreover, in our experiment, count of n-gram performs worse than the majority baseline.

%{\bf N-gram Features:}
Following prior work, we count each unigram across, as well as unique to, S1 or S2~\cite{daxenbergerG13g,zhang2015l}. However, we also count bigrams and trigrams to better capture introduced or deleted argumentative discourse units.

%{\bf Textual Features:}
Another group of features are based on sentence differences similar to those proposed in~\cite{zhang2015l}, e.g., difference in length, commas, symbols, named entities, etc., as well as edit distance. However, to capture improvement rather than just difference, we also introduce asymmetric distance metrics, e.g. Kullback-Leibler divergence\footnote{Using scipy.stats.entropy on sentence vectors.}.  We also capture differences using  BLEU\footnote{Using \textit{sentence\_bleu} from nltk.translate.bleu\_score module, with S1 as reference and S2 as hypothesis.} score, %(bilingual evaluation understudy)  
motivated by its use in evaluating machine-translated text quality.

%{\bf Language Features:}
Following~\newcite{zhang2015l}, we calculate the count and difference of spelling and language errors\footnote{Using python `language-check' tool.}, in our case to capture improvement as a result of error corrections. 

As stated in the annotation guidelines, one way a revised sentence can be better is because it is  more precise or specific. Therefore, we introduce the use of the Speciteller~\cite{li2015specificity} tool to quantify the specificity of S1 and S2,  and take the specificity difference as a new feature.

%\paragraph{Parse Tree Features:}
\newcite{remse2016msAESW}  used parse tree based features to capture the readability, coherence, and fluency of a sentence. Inspired by them, we calculate the difference in count of subordinate clauses (SBAR), verb phrases (VP), noun phrases (NP), and tree height in the parse trees\footnote{https://nlp.stanford.edu/software/lex-parser.shtml} of S1 and S2.

%%%%%%%%%%%%%%%%%%%%%%%%%%%%%%%%%%%%%%%%%%%%%%%%%%%%%%%%%%%%%%%%%%%%%%%%%%%%%%%%
\section{Experiments and Results}
\label{sec: experiments}
Our goal is to examine whether we can predict improvement for non-expert ArgRewrite revisions, using AESW expert and/or  ArgRewrite non-expert revisions for training. Our experiments are structured to answer the following research questions:

\textbf{Q1:} Can we use only non-expert revisions to train a model that outperforms a baseline?

\textbf{Q2:} Can we use only expert revisions to train a model that outperforms a   baseline?

\textbf{Q3:} Can we combine expert and non-expert training revisions to improve model performance?

%\subsection{Experimental Setup}
%To answer the questions we ran machine learning experiments and report the results in Section~\ref{sec: results}. 
Our machine learning experiments use Random Forest (RF)~\footnote{Random Forest outperformed Support Vector Machines.} from Python scikit-learn toolkit~\cite{scikit-learn} with 10-fold cross validation. Parameters were tuned using AESW development data.
Because of the ArgRewrite class imbalance  (Table~\ref{table:inter-annotator agreement}, All row), we used SMOTE~\cite{chawla2002bhk} oversampling for each training fold. Feature selection was also performed on each training fold. Average un-weighted precision, recall and F1 are reported and compared to majority-class baselines.
%\footnote{An n-gram baseline~\cite{daxenbergerG13g,zhang2015l} performed worse than majority.}

%%%%%%%%%%%%%%%%%%%%%%%%%%%%%%%%%%%%%%%%%%%%%%%%%%%%%%%%%%%%%%%%%%%%%%%%%%%%%%%%

%\section{Results}
%\label{sec: results}
\begin{table}
\centering
\resizebox{\columnwidth}{!}{%
\begin{tabular}{|l|l|l|l|l|}
    \hline
	\textbf{Experiments}	 &	\textbf{Precision}  	& \textbf{Recall} & \textbf{F1} \\  \hline
\hline 
 Majority	baseline			&	0.417	&	0.500	&	0.454	\\ \hline	\hline	
AESW	all			&	0.471*	&	0.470	&	0.468	\\	\hline	
AESW	plaintext			&	0.511*	&	0.515	&	0.473	\\	\hline	\hline
ArgRewrite				&	0.570*	&	0.534	&	0.525*	\\	\hline	
ArgRewrite	+	AESW	all	&	0.497*	&	0.501	&	0.488*	\\	\hline	
ArgRewrite	+	AESW	plaintext	&	\textbf{0.574*}	&	\textbf{0.555*}	&	\textbf{0.551*}	\\	 \hline
 %AESW non-MATH+EditDistance[1,20] &	0.564*	 & 0.543*  & 0.542*	\\ \hline
\end{tabular}
}
\caption{10-fold cross-validation performance. * indicates significantly better than majority ($p<0.05$). Bold indicates highest column value.}
%\vspace{-1em}
\label{table: results}
\end{table}

\begin{figure}
\begin{center}
%\fbox{\parbox{6cm}{
%This is a figure with a caption.}}
\includegraphics[width=0.49\textwidth]{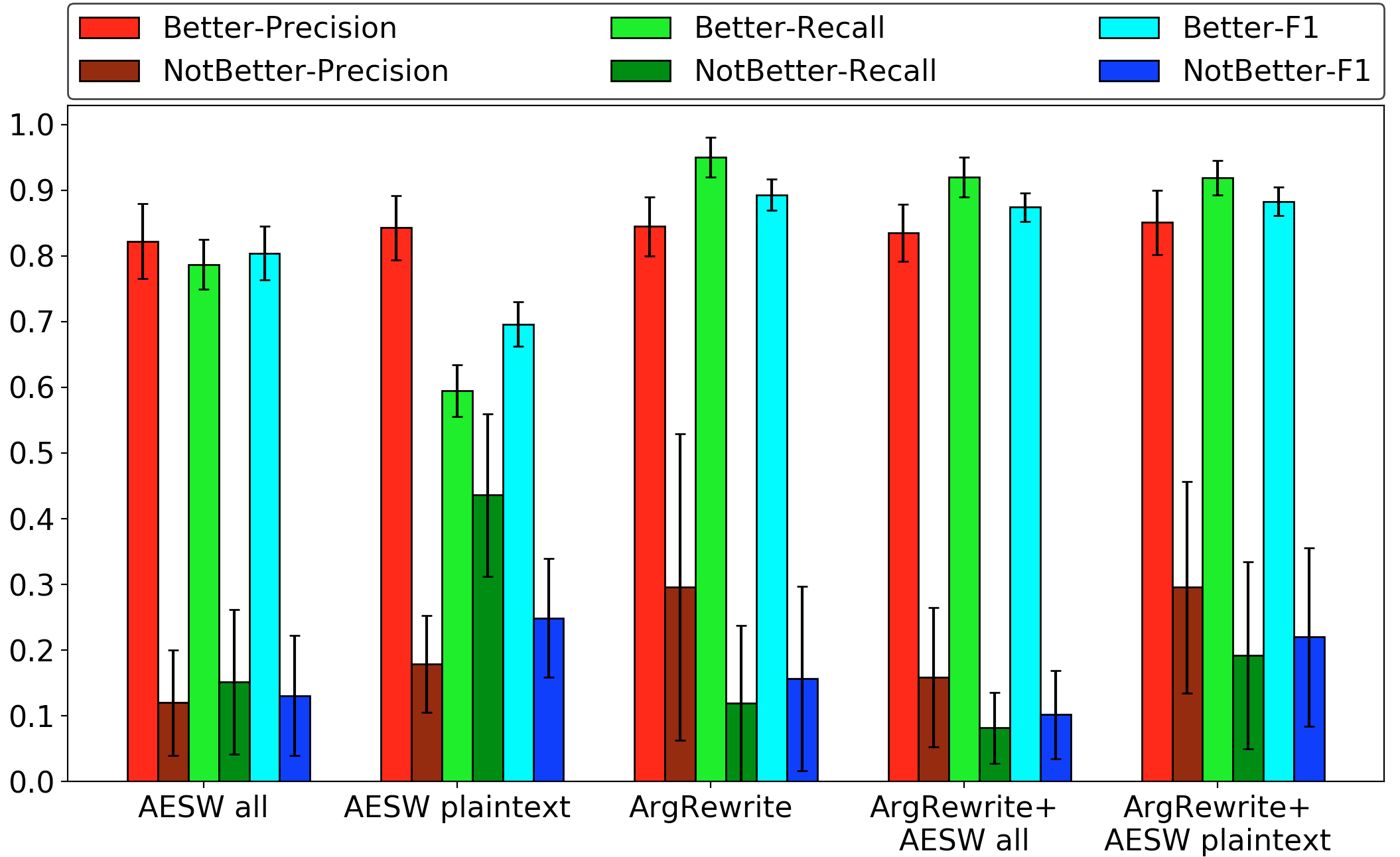} 
\caption{Precision, Recall, and F1 by class label.} 
%Orange lines shows the values for the ArgRewrite interface and blue lines shows the values for the diff based interface.}
\label{fig: prf better-notbetter}
\vspace{-1.5em}
\end{center}
\end{figure}

%%%%%%%%%%%%%%%%% Table %%%%%%%%%%%%%%%%%%%%%%%%%
\begin{table*}[t]
\centering
\resizebox{2.1\columnwidth}{!}{
\begin{tabular}
%{|p{0.6\columnwidth}|p{0.6\columnwidth}|p{0.18\columnwidth}|p{0.218\columnwidth}|p{0.9\columnwidth}|}
{|p{0.77\columnwidth}|p{0.77\columnwidth}|p{0.218\columnwidth}|p{0.9\columnwidth}|}
    \hline
    \textbf{Original Sentence ($S1$)} & \textbf{Revised Sentence ($S2$)}% & \textbf{Majority Label}  
    & \textbf{Label Distribution} & \textbf{Sample Comments}\\ 
    \hline
    
    A 1,000-word letter is considered long, and takes days, if not weeks, to reach the recipient.	
    & A 1,000-word letter is considered long, and takes days, if not weeks, to reach the recipient, with risks of getting lost along the way.	
    %& NotBetter	
    & 3 vs 4	
    & \textbf{NotBetter:} S1 is clearer than S2 and the 'risks along the way' could be included as a second sentence to increase readability.\hspace{10cm} \textbf{Better:} S2 provides more information that strengthens the idea/major claim in S1. 
    \\
    \hline
    People can't feel the atmosphere of the conversation.	
    & Also, people can't feel the atmosphere of the conversation.	
    %& NotBetter	
    & 3 vs 4	
    & \textbf{NotBetter:} Either sentence is fine, but sentence two is not any better.\hspace{10cm} \textbf{Better:} Assuming this sentence originally came from the context of a larger part of text, I imagine the continuation included here improves the flow of the original context.
    \\
    \hline
    
    With respect to personal life, social networking provides us opportunities to interact with people from different areas, such as Facebook and Twitter.
    & With respect to personal life, social networkings provide us opportunities to interact with people from different areas, such as Facebook and Twitter.
    %& NotBetter	
    & 1 vs 6	
    & \textbf{NotBetter:} S2 includes incorrect grammar.\hspace{10cm} \textbf{Better:} S1 flows better than S2.
    \\
    \hline
    
\end{tabular}
}
\caption{Misclassified $NotBetter$ revisions from ArgRewrite along with label distribution ($\#Better$ vs $\#NotBetter$) and sample annotator comments.}
\label{table:misclassified revisions}
\end{table*}

To answer Q1, we train a model using only ArgRewrite data. Table~\ref{table: results} shows that this model outperforms the majority baseline, significantly so for Precision and F1. Compared to all other models (Figure~\ref{fig: prf better-notbetter}), this model can identify `Better' revisions with the highest recall,  and can identify `NotBetter' revisions with the highest precision. However, for our long-term goal of building an effective revision assistant tool, intuitively we will also need to identify `NotBetter' revisions with higher recall, which is very low for this model.

To answer Q2, we train only on AESW data but test on the same ArgRewrite folds as above. For both AESW revision samples (before and after removing the placeholders), only Precision is significantly better than the baseline. However, Figure~\ref{fig: prf better-notbetter} shows that AESW plaintext has significantly higher $(p<0.05)$ Recall than any other model in predicting `NotBetter' revisions (which motivates Q3 as a way to address the limitation noted in Q1). %Hence, to utilize AESW data to address the limitation identified for Q1, we mix AESW with ArgRewrite when investigating Q3.

To answer Q3, during each run of cross-validation training we inject the AESW data in addition to the $90\%$ ArgRewrite data, then test on the remaining $10\%$ as before. As can be seen from Table~\ref{table: results}, 
AESW plaintext combined with ArgRewrite shows the best classification performance using all three metrics. It also has improved Recall  for `NotBetter' revisions compared to training only on ArgRewrite data. This result indicates that selective extraction of revisions from AESW data helps improve  model performance, especially when classifying low-quality revisions.

Finally, to understand feature utility, we compute  average  feature importance in the 10-folds for each experiment.
%and compare important features for each of the experiments. 
Top important features include unigrams, trigrams, length difference, language errors, edit distance, BLEU score, specificity difference, and parse-tree features.
For example, length difference  scores in the top 5 for all experiments. This is intuitive as the annotation guidelines state that adding evidence can make a better revision.
%, which creates longer sentences.  
Other features such as differences in language errors, specificity scores, and BLEU scores show more importance when training on combined ArgRewrite and AESW data than when training on only ArgRewrite. Surprisingly, spelling error corrections show low importance.  %Although we use count of POS and NE features, they do not show any influence for any of our experiment.

%In the future, we would like to investigate two directions. First, our current features are largely borrowed from prior work.  During annotation we in fact collected free text justifications for each annotated example, which we plan to use for feature development.
%Second, we would like to improve our use of the AESW data, e.g., by manually using the annotation guidelines or automatically clustering revisions for more targeted sampling. 
%would help us further improve the model performance and assist in automatic feedback about revision quality.
%Finally, although we have not analyzed the comments, we believe careful observation of the comments will give us insights for developing features for future efforts in automatic classification.

\section{Discussion} 
Although AESW-plaintext helped classify NotBetter revisions, performance is still low. 
%our models categorize Better revisions with higher accuracy compared to the NotBetter revisions. 
%We look deeply into the misclassified revisions and the collected annotator comments.% of the most difficult to classify examples. 
Table~\ref{table:misclassified revisions} shows some example NotBetter revisions  misclassified as Better by most models. The first two examples were also difficult for humans to classify.

In the first example,  one annotator for Better (the minority label) points out that the revision provides more information. %Our models have positive average length difference\footnote{Length difference = $|S2|-|S1|$} for Better revisions, which means length increase, and negative average length difference for NotBetter revisions, which means length decrease. 
We speculate that our models might similarly rely too heavily on length and classify longer sentences as Better, since as noted above, length difference was a top 5 feature in all experiments.  In fact, for the best model (ArgRewrite+AESW plaintext), the length difference for predicted Better revisions was $4.81$, while for predicted NotBetter revisions it was $-3.99$.  %The inter-rater reliability of our annotation task is low because of similar label distribution.  although the label distribution indicates low confidence on the majority label. 

In the second example, %while a majority of the annotators labeled the revision as NotBetter.
%because the added word did not improve quality or clarity of the revision. 
one of the annotators who labeled the revision as Better noted that the added word `Also' indicates a larger context not available to the annotators. This suggests that including revision context  could help improve both annotation and classification performance.

The third revision was annotated as NotBetter by $6$ annotators. We looked into our features %to see if they correctly extracted information about grammar mistakes 
and found that the `language-check' tool in fact was able to catch this grammatical mistake. %, yes most of our machine-learning models still misclassified the revision. %Even though for this particular example our features were as expected, yet we do not have the resource to manually extract all features (e.g. specificity, parse tree features etc.) for every instance to check feature accuracy. 
Yet only the model using just ArgRewrite for training was able to correctly classify this revision, as all models using AESW data misclassified. 

\section{Conclusion and Future Work}
\label{section:conclusion and future work}
We created a corpus of sentence-level student revisions annotated with labels regarding improvement with respect to argumentative writing.\footnote{Freely available for research usage at: http://www.petal.cs.pitt.edu/data.html} We used this corpus to build a machine learning model for automatically identifying revision improvement. We also demonstrated smart use of an existing corpus of expert edits to improve model performance.  

In the future, we would like to improve inter-rater reliability by collecting expert annotations rather than using crowdsourcing. We would also like to examine how the accuracy of our feature extraction algorithms impacted our feature utility results. Finally, we would like to improve our use of the AESW data, e.g., by automatically clustering revisions for more targeted sampling. Optimizing how many AESW revisions to use and how to balance labels in AESW sampling are also areas for future research.

\section*{Acknowledgments}
We thank Prof. Rebecca Hwa for her guidance in setting up a pilot predating our current annotation procedure. We would also like to thank Luca Lugini, Haoran Zhang and members of the PETAL group for their helpful suggestions and the anonymous reviewers for their valuable advice. This research is funded by NSF Award 1550635. % and 1735752.

\bibliography{naaclhlt2018}

\begin{thebibliography}{24}
\expandafter\ifx\csname natexlab\endcsname\relax\def\natexlab#1{#1}\fi

\bibitem[{Adler et~al.(2011)Adler, de~Alfaro, Mola-Velasco, Rosso, and
  West}]{adler2011dm}
B.~Thomas Adler, Luca de~Alfaro, Santiago~M. Mola-Velasco, Paolo Rosso, and
  Andrew~G. West. 2011.
\newblock Wikipedia vandalism detection: Combining natural language, metadata,
  and reputation features.
\newblock In \emph{Computational Linguistics and Intelligent Text Processing},
  CICLing '11, pages 277--288, Berlin, Heidelberg. Springer Berlin Heidelberg.

\bibitem[{Bronner and Monz(2012)}]{bronner2012m}
Amit Bronner and Christof Monz. 2012.
\newblock \href {http://dl.acm.org/citation.cfm?id=2380816.2380860} {User edits
  classification using document revision histories}.
\newblock In \emph{Proceedings of the 13th Conference of the European Chapter
  of the Association for Computational Linguistics}, EACL '12, pages 356--366,
  Avignon, France. Association for Computational Linguistics.

\bibitem[{Chawla et~al.(2002)Chawla, Bowyer, Hall, and
  Kegelmeyer}]{chawla2002bhk}
Nitesh~V. Chawla, Kevin~W. Bowyer, Lawrence~O. Hall, and W.~Philip Kegelmeyer.
  2002.
\newblock Smote: Synthetic minority over-sampling technique.
\newblock \emph{Journal of Artificial Intelligence Research}, 16(1):321--357.

\bibitem[{Dahlmeier and Ng(2012)}]{dahlmeier2012n}
Daniel Dahlmeier and Hwee~Tou Ng. 2012.
\newblock \href {http://dl.acm.org/citation.cfm?id=2382029.2382118} {Better
  evaluation for grammatical error correction}.
\newblock In \emph{Proceedings of the 2012 Conference of the North American
  Chapter of the Association for Computational Linguistics: Human Language
  Technologies}, NAACL HLT '12, pages 568--572, Stroudsburg, PA, USA.
  Association for Computational Linguistics.

\bibitem[{Daudaravicius et~al.(2016)Daudaravicius, Banchs, Volodina, and
  Napoles}]{daudaravicius2016bv}
Vidas Daudaravicius, Rafael~E. Banchs, Elena Volodina, and Courtney Napoles.
  2016.
\newblock \href {http://aclweb.org/anthology/W/W16/W16-0506.pdf} {A report on
  the automatic evaluation of scientific writing shared task}.
\newblock In \emph{Proceedings of the 11th Workshop on Innovative Use of {NLP}
  for Building Educational Applications, BEA@NAACL-HLT 2016, June 16, 2016, San
  Diego, California, {USA}}, pages 53--62.

\bibitem[{Daxenberger and Gurevych(2012)}]{daxenberger2012g}
Johannes Daxenberger and Iryna Gurevych. 2012.
\newblock A corpus-based study of edit categories in featured and non-featured
  wikipedia articles.
\newblock In \emph{Proceedings of the 24th International Conference on
  Computational Linguistics}, COLING '12, pages 711--726, Mumbai, India.

\bibitem[{Daxenberger and Gurevych(2013)}]{daxenbergerG13g}
Johannes Daxenberger and Iryna Gurevych. 2013.
\newblock \href {http://aclweb.org/anthology/D/D13/D13-1055.pdf} {Automatically
  classifying edit categories in wikipedia revisions}.
\newblock In \emph{Proceedings of the 2013 Conference on Empirical Methods in
  Natural Language Processing}, EMNLP '13, pages 578--589, Seattle, Washington,
  USA. Association for Computational Linguistics.

\bibitem[{Eli~Review(2014)}]{elireview}
The Eli~Review. 2014.
\newblock \href {https://elireview.com/} {Eli review, https://elireview.
  com/support/. [online; accessed 03-18-2018].}

\bibitem[{Felice et~al.(2016)Felice, Bryant, and Briscoe}]{felice2016bb}
Mariano Felice, Christopher Bryant, and Ted Briscoe. 2016.
\newblock Automatic extraction of learner errors in esl sentences using
  linguistically enhanced alignments.
\newblock In \emph{Proceedings of COLING 2016, the 26th International
  Conference on Computational Linguistics: Technical Papers}, pages 825--835.

\bibitem[{Fleiss(1971)}]{fleiss1971}
Joseph~L. Fleiss. 1971.
\newblock {Measuring nominal scale agreement among many raters}.
\newblock \emph{Psychological Bulletin}, 76(5):378--382.

\bibitem[{Grammarly(2016)}]{grammarly}
Grammarly. 2016.
\newblock http://www.grammarly.com. [online; accessed 03-18-2018].

\bibitem[{Javanmardi et~al.(2011)Javanmardi, McDonald, and
  Lopes}]{javanmardi2011ml}
Sara Javanmardi, David~W. McDonald, and Cristina~V. Lopes. 2011.
\newblock Vandalism detection in wikipedia: A high-performing, feature-rich
  model and its reduction through lasso.
\newblock In \emph{Proceedings of the 7th International Symposium on Wikis and
  Open Collaboration}, WikiSym '11, pages 82--90, New York, NY, USA. ACM.

\bibitem[{Landis and Koch(1977)}]{landis77}
J.~Richard Landis and Gary Koch. 1977.
\newblock The measurement of observer agreement for categorical data.
\newblock \emph{Biometrics}, 33(1):159--174.

\bibitem[{Li and Nenkova(2015)}]{li2015specificity}
Junyi~Jessy Li and Ani Nenkova. 2015.
\newblock Fast and accurate prediction of sentence specificity.
\newblock In \emph{Proceedings of the Twenty-Ninth Conference on Artificial
  Intelligence (AAAI)}, pages 2281--2287.

\bibitem[{Paulus(1999)}]{paulus1999}
Trena~M. Paulus. 1999.
\newblock The effect of peer and teacher feedback on student writing.
\newblock \emph{Journal of Second Language Writing}, 8(3):265 -- 289.

\bibitem[{Pedregosa et~al.(2011)Pedregosa, Varoquaux, Gramfort, Michel,
  Thirion, Grisel, Blondel, Prettenhofer, Weiss, Dubourg, Vanderplas, Passos,
  Cournapeau, Brucher, Perrot, and Duchesnay}]{scikit-learn}
F.~Pedregosa, G.~Varoquaux, A.~Gramfort, V.~Michel, B.~Thirion, O.~Grisel,
  M.~Blondel, P.~Prettenhofer, R.~Weiss, V.~Dubourg, J.~Vanderplas, A.~Passos,
  D.~Cournapeau, M.~Brucher, M.~Perrot, and E.~Duchesnay. 2011.
\newblock Scikit-learn: Machine learning in {P}ython.
\newblock \emph{Journal of Machine Learning Research}, 12:2825--2830.

\bibitem[{Remse et~al.(2016)Remse, Mesgar, and Strube}]{remse2016msAESW}
Madeline Remse, Mohsen Mesgar, and Michael Strube. 2016.
\newblock \href
  {http://aclanthology.coli.uni-saarland.de/pdf/W/W16/W16-0518.pdf}
  {Feature-rich error detection in scientific writing using logistic
  regression}.
\newblock In \emph{Proceedings of the 11th Workshop on Innovative Use of {NLP}
  for Building Educational Applications, BEA@NAACL-HLT 2016, June 16, 2016, San
  Diego, California, {USA}}, pages 162--171. Association for Computational
  Linguistics.

\bibitem[{Tan and Lee(2014)}]{tan2014l}
Chenhao Tan and Lillian Lee. 2014.
\newblock A corpus of sentence-level revisions in academic writing: {A} step
  towards understanding statement strength in communication.
\newblock In \emph{Proceedings of the 52nd Annual Meeting of the Association
  for Computational Linguistics}, volume 2: Short Papers, pages 403--408,
  Baltimore, MD, USA.

\bibitem[{Turnitin(2014)}]{turnitin}
Turnitin. 2014.
\newblock \href {http://turnitin.com/} {http://turnitin.com/. [online; accessed
  03-18-2018].}

\bibitem[{Writing~Mentor(2016)}]{ets-writing-mentor}
The Writing~Mentor. 2016.
\newblock \href {https://mentormywriting.org/} {Ets writing mentor,
  https://mentormywriting.org/, [online; accessed 03-18-2018].}

\bibitem[{Xue and Hwa(2014)}]{xue2014h}
Huichao Xue and Rebecca Hwa. 2014.
\newblock \href {http://www.aclweb.org/anthology/P14-2098} {Improved correction
  detection in revised esl sentences}.
\newblock In \emph{Proceedings of the 52nd Annual Meeting of the Association
  for Computational Linguistics (Volume 2: Short Papers)}, pages 599--604,
  Baltimore, Maryland. Association for Computational Linguistics.

\bibitem[{Yang et~al.(2017)Yang, Halfaker, Kraut, and Hovy}]{yang2017hkh}
Diyi Yang, Aaron Halfaker, Robert~E. Kraut, and Eduard~H. Hovy. 2017.
\newblock Identifying semantic edit intentions from revisions in wikipedia.
\newblock In \emph{Proceedings of the 2017 Conference on Empirical Methods in
  Natural Language Processing, {EMNLP} 2017, Copenhagen, Denmark, September
  9-11, 2017}, pages 2000--2010. Association for Computational Linguistics.

\bibitem[{Zhang et~al.(2017)Zhang, Hashemi, Hwa, and Litman}]{zhang2017hh}
Fan Zhang, Homa Hashemi, Rebecca Hwa, and Diane Litman. 2017.
\newblock \href {https://doi.org/10.18653/v1/P17-1144} {A corpus of annotated
  revisions for studying argumentative writing}.
\newblock In \emph{Proceedings of the 55th Annual Meeting of the Association
  for Computational Linguistics (Volume 1: Long Papers)}, pages 1568--1578.
  Association for Computational Linguistics.

\bibitem[{Zhang and Litman(2015)}]{zhang2015l}
Fan Zhang and Diane Litman. 2015.
\newblock Annotation and classification of argumentative writing revisions.
\newblock In \emph{Proceedings of the 10th Workshop on Innovative Use of NLP
  for Building Educational Applications}, pages 133--143, Denver, Colorado.
  Association for Computational Linguistics.

\end{thebibliography}
\bibliographystyle{acl_natbib}

\end{document}